\documentclass[fleqn,10pt]{wlscirep}
\usepackage[utf8]{inputenc}
\usepackage[T1]{fontenc}
\usepackage{listings}
\usepackage{cleveref}

\newcommand{\beginsupplement}{%
  \setcounter{figure}{0}\renewcommand{\thefigure}{S\arabic{figure}}%
  \setcounter{table}{0}\renewcommand{\thetable}{S\arabic{table}}%
  \setcounter{equation}{0}\renewcommand{\theequation}{S\arabic{equation}}%
}

\newif\ifsupp
\supptrue    

\newif\ifstatements
\statementsfalse    

\title{An end-to-end-trained vision–language model for native-language prostate pathology report generation}

\author[1,2,3*]{Christian Grashei}
\author[1,2,3]{Fabian Gülhan}
\author[4]{Maximilian Legnar}
\author[1]{Fabian Stögbauer}
\author[4,5]{Cleo-Aron Weis}
\author[1]{Carolin Mogler}
\author[1,2,3]{Peter Schüffler}
\affil[1]{Institute of Pathology, Technical University of Munich, Munich, Germany}
\affil[2]{Munich Data Science Institute, Munich, Germany}
\affil[3]{Munich Center for Machine Learning, Munich, Germany}
\affil[4]{Section Computational Pathology Heidelberg, Institute of Pathology Heidelberg, University Hospital Heidelberg, Heidelberg, Germany}
\affil[5]{Interdisciplinary Center for Scientific Computing (IWR), Heidelberg University, Heidelberg, Germany}

\affil[*]{Corresponding Author. E-mail: christian.grashei@tum.de}

\begin{abstract}
Prostate cancer is among the most frequently diagnosed malignancies worldwide, and structured reporting of each biopsy core burdens pathologists. Existing tools frame this as classification, leaving pathologists to assemble coherent reports, while many slide-level vision-language models rely on English-centric encoders that transfer poorly to other clinical languages. We present a slide-level framework generating prostate biopsy reports that is language-independent by construction: tokenizer and model are trained from scratch, demonstrated here in German. To address paired-data scarcity, an automated pipeline uses a locally deployed large language model to split composite reports into core-specific image–text pairs, yielding 17,344 pairs from 2,402 historical cases without manual annotation. Evaluated for clinical attributes rather than linguistic similarity, the model achieves 96.2\% F1 for malignancy detection and 65.2\% for Gleason grading, competitive with an FDA-cleared classifier. Grading is further validated on three external cohorts with latent-space augmentation. Institutions can thus train native-language reporting models on their own archives.
\end{abstract}
\begin{document}

\flushbottom
\maketitle

\thispagestyle{empty}

\section*{Introduction}

Prostate cancer remains a substantial global health challenge. With over 300,000 new cases in the US alone in 2025, it accounts for approximately 30\% of all new male cancer diagnoses~\cite{siegel2025cancer}. This high disease burden imposes a substantial workload on pathology departments~\cite{strom2020artificial}.
Prostate cancer diagnosis follows a clinical pathway that typically begins with prostate-specific antigen (PSA) screening and magnetic resonance imaging (MRI) and is confirmed with a prostate core needle biopsy as the gold standard. Typically, approximately twelve biopsy cores are taken per patient. Each biopsy core must be individually reviewed to determine malignancy, assign a Gleason grade, and measure tumor volume~\cite{van20202019}. Consequently, a written pathology report documenting the findings for each biopsy core is issued to the treating physician~\cite{egevad2019dataset}. The sheer volume of these cases, combined with their highly structured reporting requirements and pattern-based Gleason grading, makes prostate pathology an ideal candidate for AI-driven transformation~\cite{marletta2024artificial}.

Previous computational approaches have largely framed prostate cancer detection as a classification problem~\cite{lucas2019deep, madabhushi2020deep}. While early patch-level methodologies achieved high accuracy, they required labor-intensive, granular annotations by trained pathologists~\cite{arvaniti2018automated}. More recent advances utilize Multiple Instance Learning (MIL)~\cite{ilse2018attention} frameworks to process whole slide images (WSIs) by aggregating embeddings from pathology-specific foundation models, outputting a class probability (e.g., benign or a specific Gleason grade)~\cite{campanella2019clinical}. However, while current AI models can assist with discrete diagnostic tasks, the cognitive work of integrating these outputs into a coherent, clinical report remains the responsibility of the pathologist.

Vision-language models (VLMs) offer a more integrated solution by directly generating descriptive text for a given tissue image~\cite{lu2024multimodal, li2023llava}. While models like PLIP~\cite{huang2023visual} and CONCH~\cite{lu2024visual} excel at tile-level captioning, they lack the WSI-level context necessary to describe an entire case. Aggregating several regions of interest~\cite{tan2024clinical} widens this context, yet still conditions the report on a subset of the slide rather than the entire WSI. HistoGPT~\cite{tran2025generating} bridges this gap in dermatopathology by generating reports directly from full WSIs, but it relies on a frozen, pre-trained language encoder (BioGPT~\cite{luo2022biogpt}). Because such encoders are predominantly English-centric, they are not applicable to non-English environments or need translation that introduces external dependencies and potential translation errors.

In this work, we present a slide-level, end-to-end report generation framework for prostate biopsies that requires no pre-trained language components. Training both a domain-specific tokenizer and the vision-language model from scratch makes the approach language-agnostic: reports are processed and generated natively in the target language, demonstrated here for German, without translation or dependence on English-centric encoders. To address the scarcity of paired WSI-text data required to develop our model, we introduce an automated pipeline that uses a locally deployed LLM to disaggregate composite reports from the laboratory information system (LIS) into core-specific WSI-text pairs, yielding 17,344 pairs from 2,402 historical cases without manual annotation.

Syntactic similarity is a poor proxy for diagnostic correctness~\cite{boag2020baselines, babar2021evaluating}, since a single altered number can potentially change a patient's management. Therefore, we depart from standard NLP metrics such as BLEU~\cite{papineni2002bleu} and ROUGE~\cite{lin2004rouge} and evaluate generated reports on the clinical attributes required by reporting guidelines: malignancy, Gleason grading, and tumor infiltration. We evaluate the model on an internal test cohort and three external cohorts, compare it against a directly supervised classification baseline and an FDA-cleared diagnostic tool~\cite{FDA2021PaigeProstate}, and introduce a latent-space augmentation strategy to mitigate scanner-induced domain shift. To our knowledge, this is the first slide-level prostate biopsy report generation model evaluated on external cohorts, and it shows that clinically useful, native-language report generation can be trained end-to-end from uncurated historical archives at a single institution.

\section*{Results}

\subsection*{An automatically curated dataset of specimen-level WSI-text pairs}
\begin{figure}[t]
\centering
\includegraphics[width=0.9\linewidth]{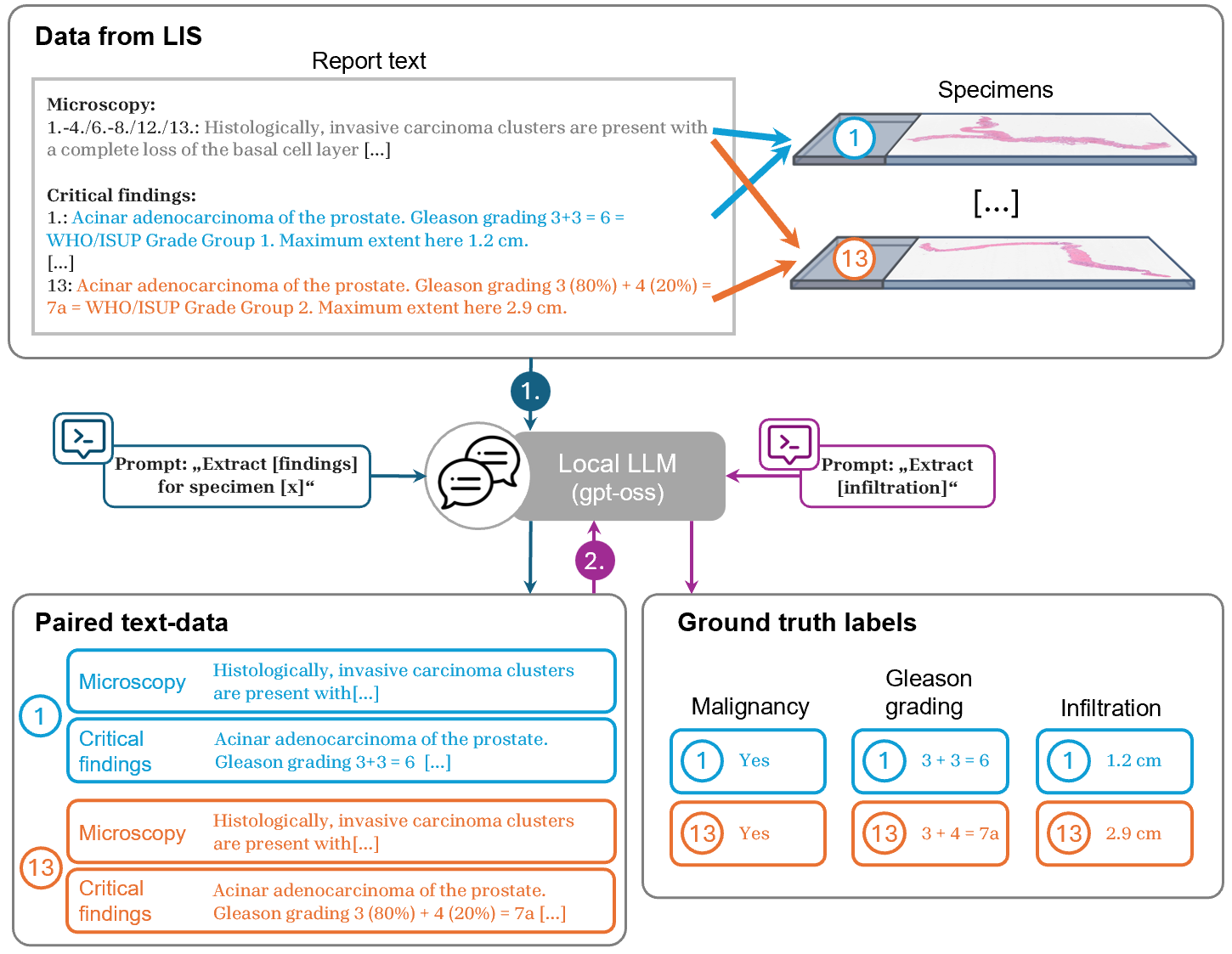}
\caption{\textbf{Generation of specimen-level image-text pairs from pathology reports using LLMs.} For a given clinical case, the complete pathology report and its associated specimen labels are retrieved from the laboratory information system (LIS). A locally executed LLM is prompted in a few-shot setting to extract information for a specific specimen from a targeted section of the report (1.). This process yields WSI-text pairs, matching the extracted report text to its corresponding WSI. Text applicable to multiple specimens is assigned to all relevant WSIs. Finally, these extracted texts are queried for the desired attributes to establish specimen-level ground truth for evaluation (2.). }
\label{fig:llm_extraction}
\end{figure}

Clinical pathology reports aggregate findings across all specimens of a case into a single document, whereas training a slide-based multimodal report generation model requires discrete pairs of WSIs and specimen-specific text. We therefore developed an automated pipeline that disaggregates composite reports into core-specific microscopy and critical findings sections using a locally deployed large language model (LLM) (Fig.~\ref{fig:llm_extraction}, see Methods).

At the Institute of Pathology of the Technical University of Munich (Munich Cohort) the pipeline yielded 17,344 WSI-text pairs derived from 2,402 unique cases, with a median text length of 47 words across the combined microscopy and critical findings sections. Text density varied systematically with diagnosis: benign cores frequently lacked detailed microscopy descriptions, whereas malignant cores yielded more granular narratives. For each core, we additionally extracted the structured labels malignancy status (benign vs.\ malignant), Gleason grade, and infiltration in percentage or extent, which enable the diagnosis-oriented evaluation reported below rather than an assessment based on linguistic similarity.

To quantify extraction accuracy, we manually audited 200 randomly sampled gradings against the original records in the Laboratory Information System (LIS). No discrepancies were observed in either the disaggregated text or the structured labels. This near-perfect extraction accuracy is consistent with prior work~\cite{lammert2025large}. The prompts used for the extraction are given in the supplementary material in Listings 1 to 3.

\subsection*{Domain-specific tokenization and native-language encoding}

The language and terminology used for pathology reports is highly domain specific. To capture these nuances, recent slide-based vision-language models~\cite{shaikovski2024prism,tran2025generating, ding2025multimodal} have utilized domain-specific LLMs such as BioGPT as partially frozen encoders. However, because these models are predominantly pre-trained on English-language corpora like PubMed, they are not suited for non-English clinical environments.

Because our clinical reports are authored natively in German, utilizing English-centric encoders presents a substantial challenge. Common workarounds, such as automated translation, introduce external dependencies and the risk of semantic drift or clinical inaccuracies. Furthermore, the vocabulary of English-trained tokenizers is poorly suited for German medical compounds. For instance, complex terms are often fragmented into numerous sub-word units, which artificially inflates sequence length and dilutes semantic density.

\begin{figure}[ht]
\centering
\includegraphics[width=0.95\linewidth]{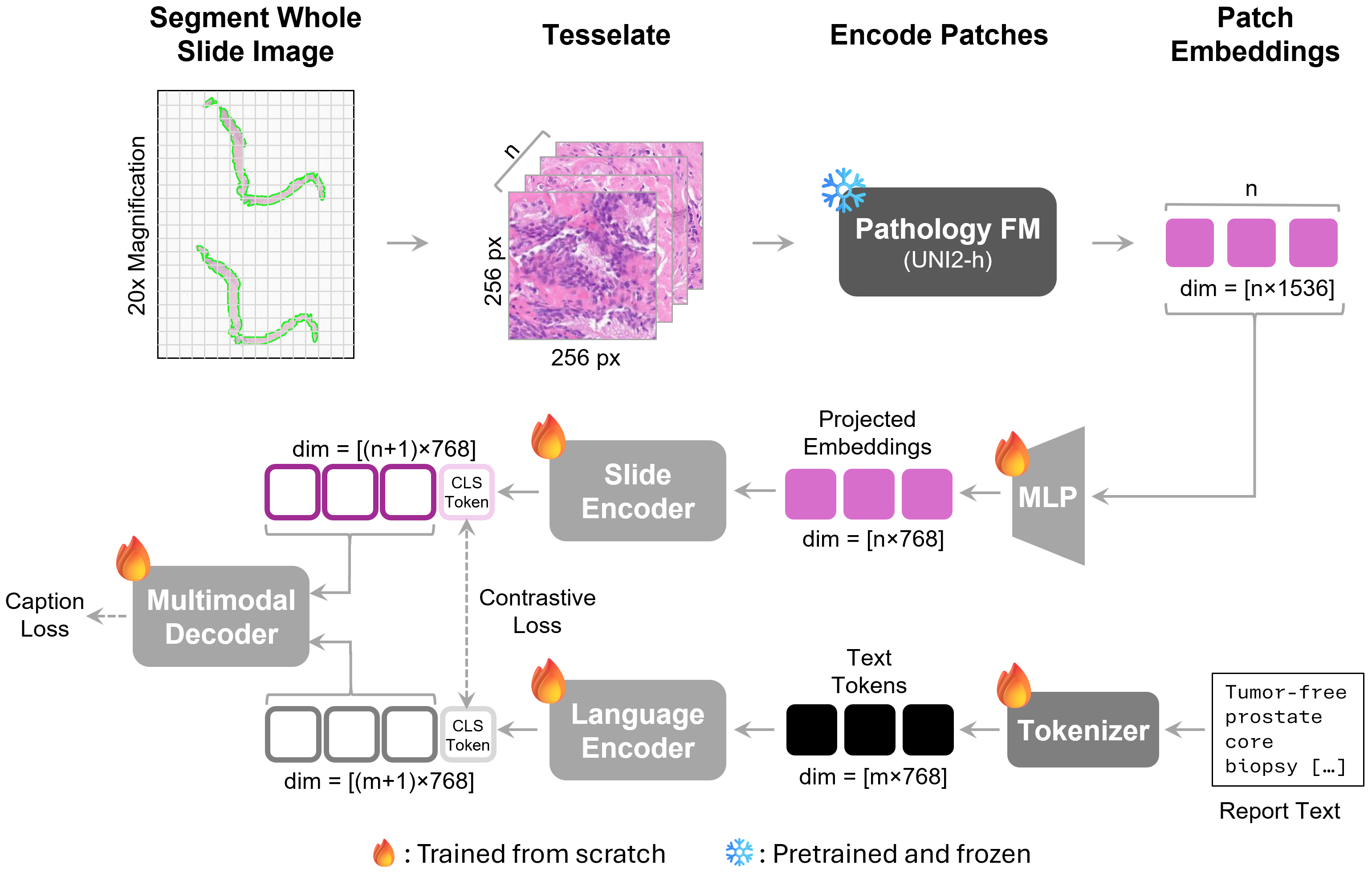}
\caption{\textbf{Overview of the preprocessing pipeline and model architecture.} The WSI is segmented to isolate the tissue area and tessellated into patches at a resolution of $256\times256$ pixels. A pathology foundation model then encodes all patches, yielding one embedding per patch. All patch embeddings for a WSI, along with the corresponding report text for the WSI, serve as input for the report generation model during training. The report model architecture comprises a projection MLP, a tokenizer, a slide encoder, a language encoder, and a multimodal decoder. The overall training objective combines a contrastive loss, computed between the outputs of the slide and language encoders, and a captioning loss applied to the report text predicted by the multimodal decoder.}
\label{fig:architecture}
\end{figure}

To overcome these limitations, a custom text tokenizer was trained exclusively on our German pathology text corpus that consists of all text samples from our extracted WSI-texts. This ensures that domain-specific German terms are represented as single or near-single tokens, substantially reducing the input vocabulary size of the tokenizer and thereby feature space. While the English-centric BioGPT utilizes a vocabulary of 42,384 tokens, our optimized tokenizer achieves higher representational efficiency with a vocabulary of only 10,142 tokens. Including the 8 mandatory special tokens, the median token number per tokenized text is 95. In conjunction with the median word count ($n=47)$, this confirms an efficient tokenization. The architecture of the model including preprocessing is shown in Fig.~\ref{fig:architecture} and explained in detail in the methods section.

\subsection*{Reporting of malignancy and grading}

Grading for prostate biopsies is done by assigning a Gleason score. The Gleason score is determined by the most common (primary) Gleason pattern and the second most common (secondary) Gleason pattern and can range from 6 to 10. Gleason scores below 6 are obsolete and are considered as benign. The information on malignancy and grading is extracted from the generated reports and compared with the extracted ground truth from the original reports.

The test dataset of the internal Munich cohort comprised 2,368 biopsies from 202 cases, isolated entirely from the 14,976 biopsies (2,200 cases) used for training the report generation model (Fig.~\ref{fig:result_internal}a). The data splits were stratified at the case level to prevent data leakage.

The report generation model correctly detected malignancy in 97.6\% of the core biopsies in the test cohort (Fig.~\ref{fig:result_internal}b). Accounting for dataset imbalance, the model achieved a specificity of 98.2\%, a sensitivity of 96.4\%, and a precision of 95.9\%, leading to an F1-score of 96.2\%. These results highlight that the model captures the morphologic cues to determine malignancy at a high accuracy.

The predicted grading was correct for 87.2\% of the biopsies in the Munich test cohort. We divided the possible grades into benign and the Gleason scores 6, 7a, 7b, 8 and 9. These classes correspond directly to the grade groups 1-5 of the International Society of Urological Pathology (ISUP)~\cite{epstein2016contemporary}. Gleason score 7a corresponds to a Gleason pattern 3 for the primary pattern and 4 for the secondary pattern, while the order is reversed for Gleason score 7b. Gleason scores 9 and 10 were merged to a single entity hereinafter because there are no samples with Gleason score 10 in the Munich dataset and it results in the same grading group according to the ISUP. The report generation model achieved a macro precision of 65.8\% and a macro recall of 65.7\% on our evaluation dataset resulting in an macro F1-score of 65.2\% (Fig. \ref{fig:result_internal}b).

An analysis of the confusion matrix reveals that the vast majority of grading errors were adjacent grade misclassifications, i.e., off by exactly one grade step (Fig.~\ref{fig:result_internal}c). This phenomenon aligns closely with known inter-observer variability among human pathologists in prostate grading~\cite{sooriakumaran2005gleason,cookson1997correlation}.
Our report texts are written by board-certified pathologists and quality-checked by a second pathologist. However, our dataset spans a time range of a decade and was created by several pathologists which might introduce variability in the grading information.

The lowest per grade recall is seen for Gleason scores 6 and 9. The higher confusion of Gleason score 9 is expected given the scarcity of samples available in our training dataset. Moreover, to distinguish Gleason scores 8 and 9 the presence of Gleason pattern 5 needs to be recognized. Among biopsies with Gleason score 9 in the training set, Gleason pattern 5 comprises an average of only 16\%, with the remainder consisting mostly of pattern 4. This underscores the difficulty of learning to distinguish these grades.


\subsection*{Reporting tumor infiltration}

Tumor infiltration quantifies the proportion of tissue occupied by cancer cells.
Because the measurement protocol changed over the time span our internal dataset covers, infiltration is recorded either as an extent (in millimeters) or as a percentage. The two formats can be treated as equivalent for clinical purposes.

\begin{figure}[htbp]
\centering
\includegraphics[width=1.0\linewidth]{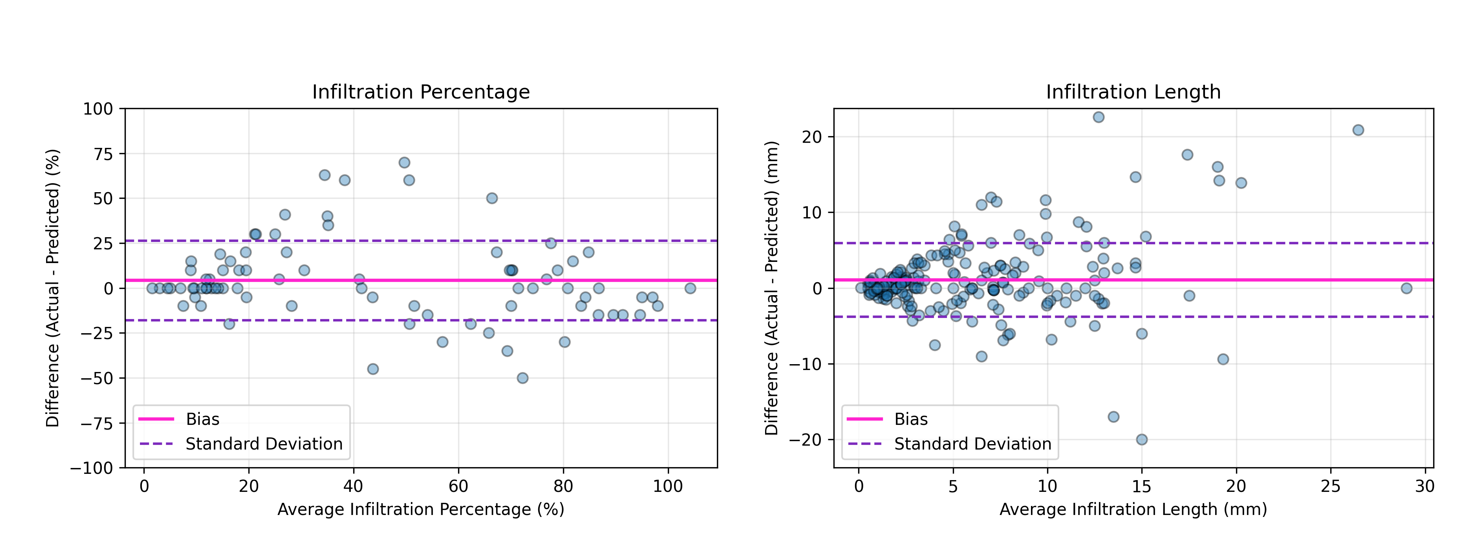}
\caption{\textbf{Evaluation of predicted infiltration in the generated reports.} Depending on the measurement protocol, the infiltration per prostate biopsy is stated as either a percentage or a physical length. Overall, the generated reports provide an accurate estimation of tumor infiltration. However, prediction deviation noticeably increases for infiltration value ranges that are underrepresented in data.}
\label{fig:result_infiltration}
\end{figure}

The output formats of ground truth and predictions were matched resulting in an evaluation cohort of 211 samples for extent (mm) and 82 samples for percentage. Both correlations were statistically significant. For the predicted infiltration in millimeters a median absolute error of 1.4 mm and a mean absolute error of 2.9 mm were measured yielding a Pearson correlation of 0.637 ($p<0.01$) and a $R^2$ score of 0.343. For predicted infiltrations in percentage the median absolute error was 10.0\% and the mean absolute error was 15.2\% with a Pearson correlation of 0.78 ($p<0.01$) and a $R^2$ score of 0.469 (Fig. \ref{fig:result_infiltration}). For both measurement protocols there is a small positive bias (overestimation) of 1.1 mm (extent) and 4.2\% (percentage). Moreover, as shown in Fig. \ref{fig:result_infiltration}, errors increase in value ranges with fewer samples.

Our results show that our report generation model learns the concept of infiltration that is contained in the report text and is able to give an estimate of the infiltration. Moreover, it is able to distinguish the two different measurement protocols that have very different value ranges (0 to 100\% for percentage vs. 0 to ~20 mm for extent). The Pearson correlation indicates a slightly higher accuracy when predicting the infiltration as percentage. This can be attributed to the imbalance of the infiltration format in our training data where percentage accounts for the most malignant samples. It should be noted that the infiltration in percentage is specified based on a visual estimation by the pathologist when viewing the biopsy under the microscope. Moreover, our report generation model is able to learn information on quantification of infiltration from a complete WSI. This highlights the advantages of slide-based approaches.

\subsection*{Comparison to an ABMIL classification model}

While our report generation directly outputs text that can be used as a draft for a pathology report, it uses text-guided weakly supervised learning. When learning from the report texts it needs to learn which parts of the text are highly relevant (e.g. grading) or less relevant (e.g. stop words) or which may even be noise (e.g. organizational comments). To evaluate the efficacy of this approach, our report generation model is compared against a directly supervised Attention-Based Multiple Instance Learning (ABMIL)~\cite{ilse2018attention} classification model trained on the exact same data partitions. The output of the classification model is a probability distribution over the possible grades (including benign).

\begin{figure}[htbp]
\centering
\includegraphics[width=0.90\linewidth]{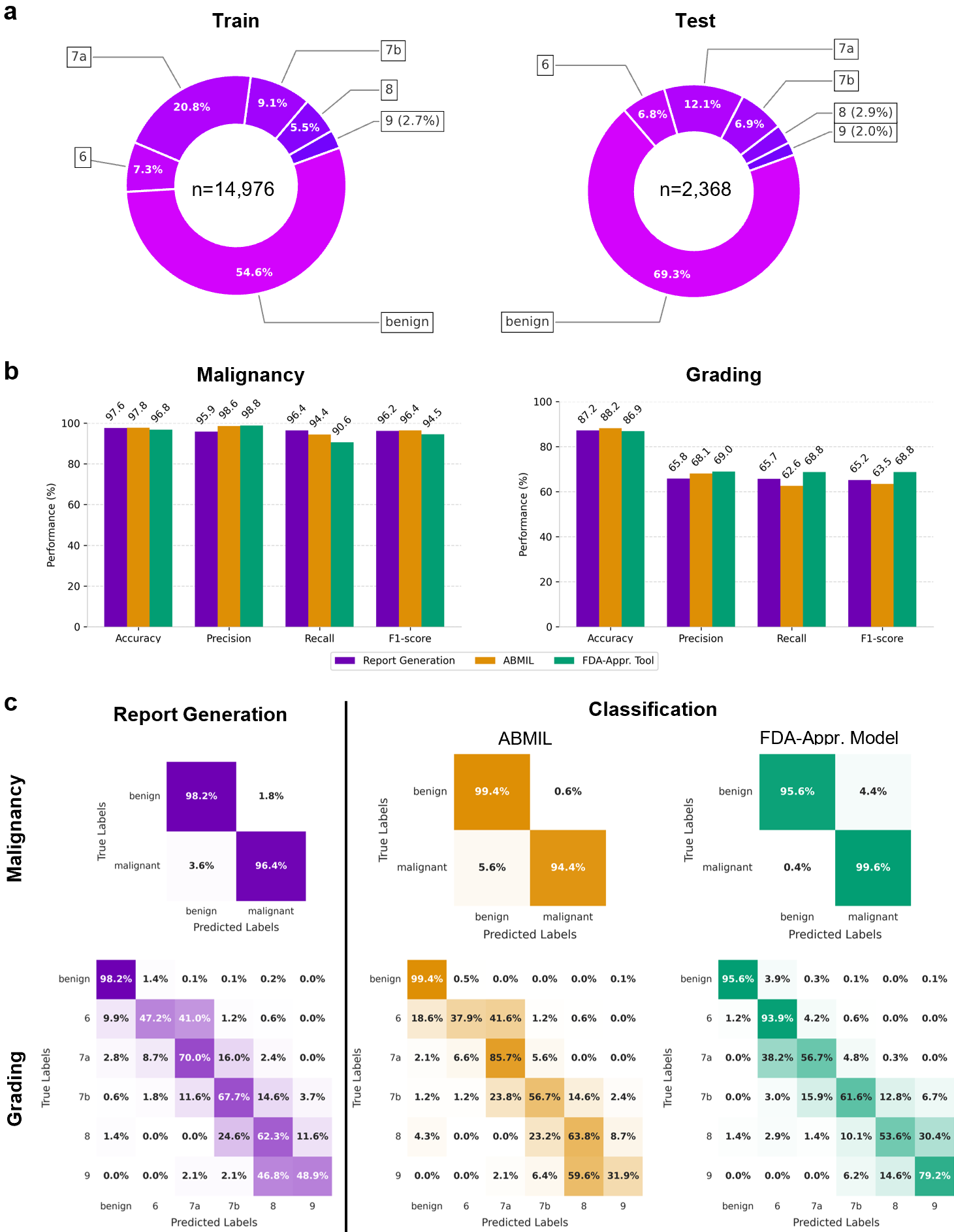}
\caption{\textbf{Train and test datasets and evaluation results on the internal Munich test cohort.}
\textbf{a} Datasets with sample size and distribution of the Gleason scores obtained during LLM extraction. 
\textbf{b} Classification metrics computed from the extracted outputs of the report generation model versus the two other models. Compared to the ABMIL model, the report generation model achieves a higher F1-score while simultaneously providing a draft report as output.
\textbf{c} Confusion matrices for predicted malignancy and grading on the Munich test dataset (rows) for our trained report generation model, an ABMIL classification model and an FDA-cleared model for prostate biopsies. Each column has the results for one model type. Report Generation Model and ABMIL model have been trained on the same data splits. The exact training dataset of the FDA-cleared model is not known.}
\label{fig:result_internal}
\end{figure}

We found that the classification model achieves a similar performance for predicting malignancy. While accuracy and precision are slightly higher for the classification model, the report generation dominates in recall leading to a difference of only 0.2\% for the F1-score between both model types (Fig. \ref{fig:result_internal}b). A similar trend in terms of precision and recall can be observed when comparing the grading performance for both model types. However, the F1-score of the report generation model is 1.7\% higher than for the classification model indicating that the report generation model is able to successfully identify the grading information in the text with high accuracy during its training.

To assess whether the two models make correlated errors, we computed Cohen's Kappa ($\kappa$) between the grading predictions of the report generation model and the classification model, and between each model's predictions and the extracted ground truth labels. Agreement between the two models ($\kappa=0.81$) is higher than the agreement of either model with the ground truth ($\kappa=0.75$ and $\kappa=0.76$), respectively. Under independent errors the reverse ordering would be expected. This indicates that the two models tend to make mistakes on the same instances.

\subsection*{Comparison to an FDA-cleared classification model}

In the recent past the first AI tools have received clearance by the U.S. Food and Drug Administration for clinical use in pathology workflows~\cite{shafi2023artificial}. To obtain clearance such tools have to undergo a rigorous evaluation to ensure their clinical suitability. To better contextualize the achieved performance of our report generation model it is compared to a commercial FDA-cleared tool (Paige Prostate) for prostate biopsies on the Munich test cohort. This FDA-cleared tool works in two stages. First, the malignancy of the biopsy core is determined. If this yields a positive result, the biopsy is graded and a Gleason score is assigned. Because the model is cloud-based and accessed via an API, the model itself is not accessible. However, we have the outputs of this tool for the Munich test cohort.

The predictions of this FDA-cleared tool for malignancy for our evaluation dataset show an accuracy of 96.8\% lagging slightly behind our report generation model. Interestingly, there is a substantial gap between precision and recall with a precision of 91.0\% and a recall of 99.6\%. It shows the model is highly sensitive to positive examples accepting false positives rather than risking a false negative that could be overlooked. Precision and recall combined lead to an F1-score of 95.1\% being 1.1\% lower compared to our model.

Evaluating the grading of this FDA-cleared tool shows an accuracy of 86.9\%, a precision of 69.0\% and a recall of 73.4\% leading to an F1-score of 68.8\%. While achieving a slightly lower accuracy in grading, the commercial tool exhibits a higher macro F1-score that takes the grade distribution of our evaluation dataset into account. Since Gleason scores $\geq9$ make up only 2.6\% of our dataset, they are massively underrepresented leading to a lower performance for this minority class for our report generation model. Nevertheless, on our internal dataset, the report generation model achieves higher recall values for the Gleason scores 7a, 7b and 8 compared to the FDA-cleared tool.

It should be noted that this commercial tool was trained on a substantially larger dataset comprising dozens of thousands of  WSIs~\cite{FDA2021PaigeProstate}. Our results show that although being trained on an uncurated, unlabeled and imbalanced dataset from a single institution our report generation model reaches performance within the same magnitude.

\subsection*{Comparison to a pan-cancer vision-language model}
While automated report generation is an active research field, a specialized, slide-based report generation model dedicated to prostate pathology has not yet been established. Existing slide-based models are primarily pan-cancer architectures trained on multi-organ pathology datasets~\cite{ding2025multimodal, sellergren2026medgemma}. A prominent, publicly available baseline is PRISM~\cite{shaikovski2024prism}, which was trained on a massive dataset comprising 587,196 slides and 195,344 reports, with prostate tissue representing a major cohort entity. Similar to HistoGPT, which is constrained to dermatopathology, PRISM utilizes BioGPT as a pretrained language encoder. 

PRISM is evaluated on the Munich test cohort using the identical evaluation protocol applied to our model. The PRISM pipeline is executed and malignancy predictions and grading metrics are extracted from the generated texts. Characteristically, the text generated by PRISM was denser (e.g., \textit{"Adenocarcinoma of prostate, Gleason score 7, Grade Group 2 identified"} or \textit{"Benign fibromuscular tissue"}). Crucially, PRISM omitted clinical information regarding the infiltration depth or percentage of the biopsy core.

In terms of tissue identification, the majority of samples were correctly classified as prostate. For 32 samples, benign biopsy cores were described as fibromuscular tissue or muscle tissue which we accepted as a benign prostate diagnosis. However, in a small subset comprising 5 samples, PRISM misidentified the prostate biopsies as esophagus, pancreas, liver, stomach, or uterus.
PRISM achieved an accuracy of 96.8\% and an F1-score of 94.5\% for predicting malignancy. Despite its use of a pretrained language encoder, its abbreviated outputs limited its clinical utility compared to our specialized model. This limitation was particularly evident in grading performance: PRISM failed to provide a grade or returned an ambiguous classification (e.g., Gleason Score 7 without a primary and secondary pattern breakdown) for 151 samples. Both types of non-specific output were retained as a separate prediction-only category. Consequently, PRISM achieved an accuracy of 80.2\%, a precision of 53.9\%, a recall of 41.7\% and an F1-score of 44.9\%. The F1-score represents a 20.3\% difference in percentage points compared to our report generation model. These results demonstrate that an accurate and clinically granular report generation model for a specific organ system can be trained effectively on a substantially smaller, domain-specific dataset ($N \approx \text{15,000}$ slides).

\begin{figure}[htbp]
\centering
\includegraphics[width=0.95\linewidth]{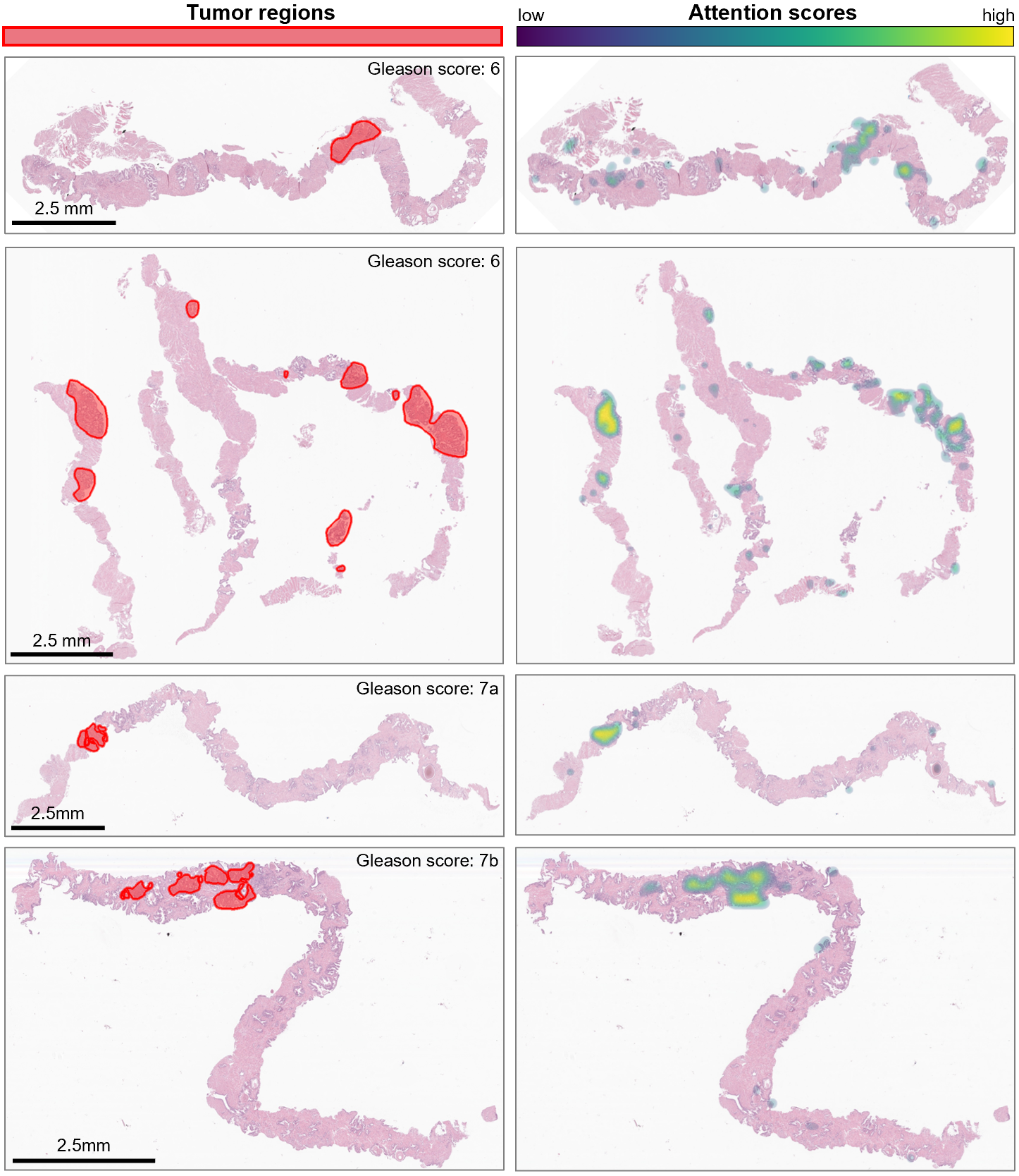}
\caption{\textbf{Visualization of tumor regions and the attention scores of the slide encoder on H\&E-stained WSIs.} Each row shows a randomly sampled malignant H\&E-stained prostate biopsy WSI from the Munich test cohort. Left: tumor regions annotated in red by a board-certified pathologist. Right: attention scores of the slide encoder of the report generation model are highlighted according to the color scale. Regions receiving high attention largely correspond with the annotated tumor areas, indicating that the slide encoder focuses on diagnostically relevant tissue and making the generated reports more interpretable.}
\label{fig:attention}
\end{figure}

\begin{figure}[t]
\centering
\includegraphics[width=0.85\linewidth]{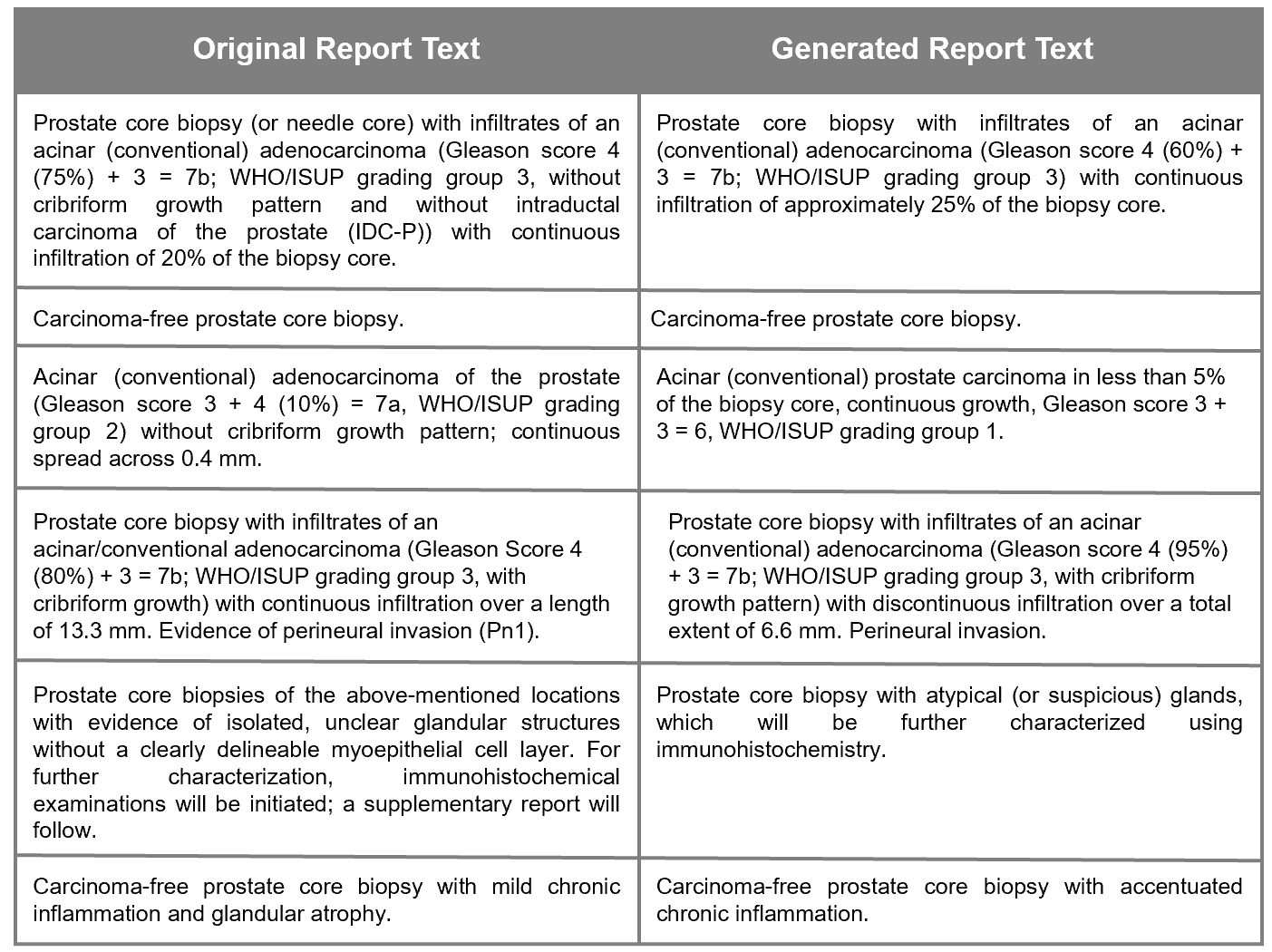}
\caption{\textbf{Examples of generated reports compared to original report texts on the Munich test cohort.} Texts are originally in German and have been translated to English for readability. The texts produced by our model can serve directly as initial drafts for clinical pathology reports. Furthermore, the model's outputs include details that enhance explainability, such as the specific proportions of Gleason grades. This additional context is particularly valuable for borderline cases. Difficult cases, where immunohistochemistry is needed for final diagnosis, may also be recognized.}
\label{fig:report_examples}
\end{figure}

\subsection*{Interpretability}
For the clinical deployment of AI models, interpretability is a crucial feature required to gain insight into the decision-making process. This transparency is important for resolving discrepancies when a generated report does not align with a pathologist's findings~\cite{tonekaboni2019clinicians}.



In Fig.~\ref{fig:attention} the tumor regions (left) and the attention weights (right) for four randomly chosen malignant samples of the Munich test cohort are visualized. Tumor regions have been annotated by a board-certified pathologist. In each of the biopsies the tumor area only makes up a fraction of the total tissue area. The attention mainly corresponds to tumor areas, allowing to trace the decision made by the report generation model. This is especially useful if the tumor only infiltrates a small area. Since the report texts used to train the model can also contain information on other conditions such as inflammation, attention may focus mainly on tumor area but not solely.

\begin{figure}[th]
\centering
\includegraphics[width=0.65\linewidth]{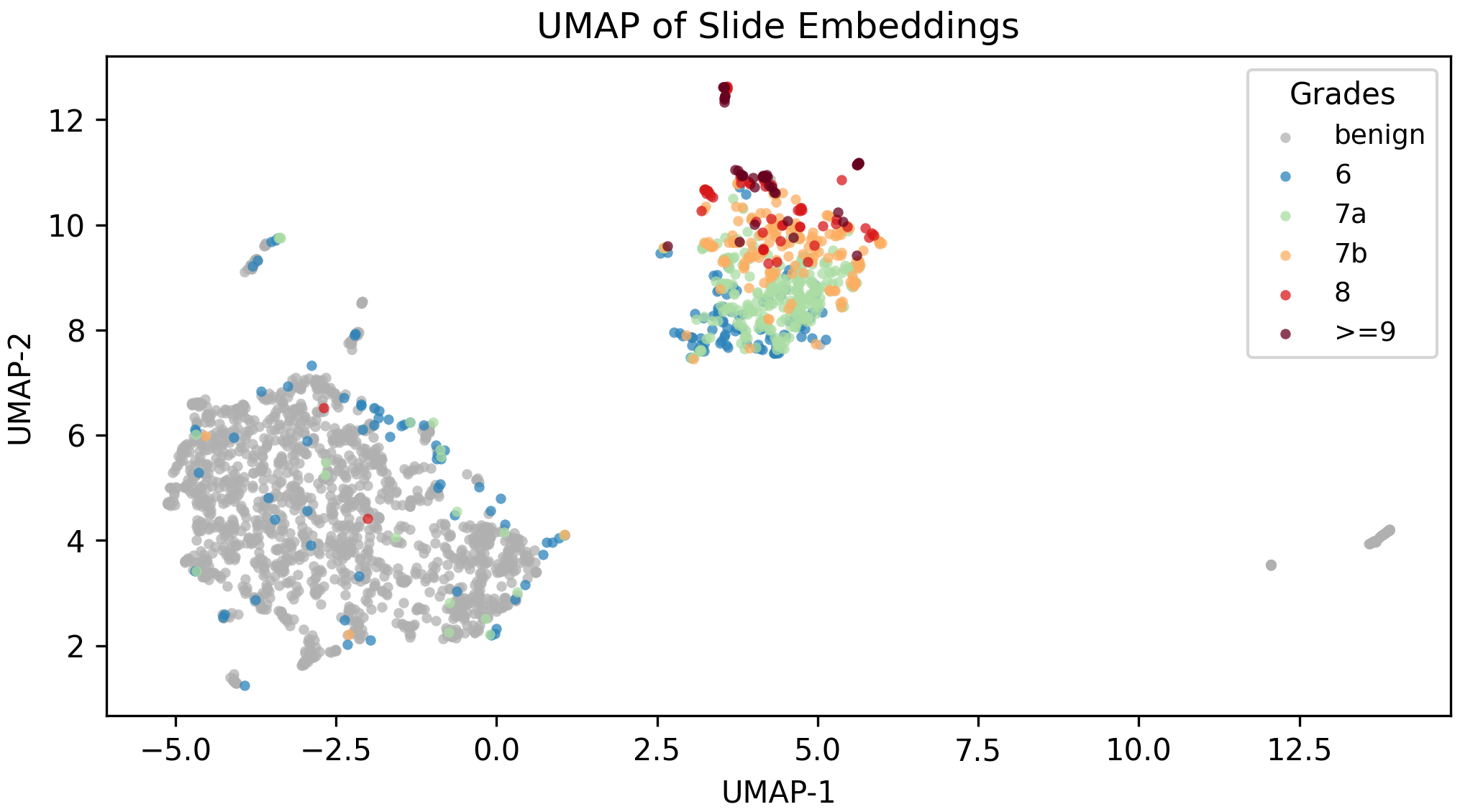}
\caption{\textbf{UMAP visualization of slide encoder embeddings.} A UMAP projection of the slide embeddings from the Munich test cohort reveals two primary clusters, corresponding to benign and malignant biopsies. Additionally, the malignant cluster demonstrates a continuous transition from lower to higher Gleason grades along the y-axis.}
\label{fig:umap}
\end{figure}

The slide embeddings that are used to compute the contrastive loss against the text embeddings can be visualized using a UMAP~\cite{mcinnes2018umap}. Fig.~\ref{fig:umap} shows a plot of the UMAP of the slide embeddings for the Munich test cohort. Colors indicate malignancy and grading information. Two larger clusters divided by a larger margin can be identified: One mainly consisting of benign samples and the other one being malignant samples. The malignant cluster further shows a transition along the y-axis. Following the y-axis upwards a transition from lower to higher grades can be seen. Furthermore, the UMAP confirms that high grades 8 and $\geq9$ are difficult to differentiate.

Further interpretability is provided through the textual outputs of the model (Fig. \ref{fig:report_examples}). Because the proportion of Gleason grades is documented in most pathology reports within the training dataset, this metric serves as an additional learning signal. Specifically, it provides explainability within the generated reports by indicating a directional tendency toward higher or lower Gleason grades—such as distinguishing between Gleason scores $3+4$ (7a) and $4+3$ (7b). For borderline or challenging cases, the report texts also contain information regarding whether immunohistochemistry (IHC) was performed. The model can leverage this pattern to incorporate IHC contexts into its output, effectively signaling to the pathologist when an IHC workup may be needed. Furthermore, the model can identify other clinically relevant conditions, such as inflammation, which may contextualize and explain an initially elevated PSA value from a prior blood sample.

\subsection*{External cohorts and mitigation of domain shift}\label{sec:augmentation}

To assess whether our trained report generation model transfers beyond the institution it was developed on, we assembled three external cohorts: a private cohort from University Hospital Heidelberg (424 biopsy cores from 50 cases) and the two cohorts of the public PANDA dataset, from Radboud University Medical Center (5,160 biopsies) and Karolinska Institutet (5,456 biopsies). All provide ground-truth malignancy and Gleason grading labels, and none contributed data to model training or hyperparameter selection. Because our model was trained on WSIs from a single institution digitized on a single scanner, we first characterized how these cohorts differ from the training distribution in color appearance. Slides from Heidelberg appear paler than those from Karolinska, which show the most intense staining of all four cohorts, while Munich and Radboud differ mainly in contrast (supplementary material Fig. S1). To quantify this, we sampled 100 WSIs at random per cohort and computed mean hue and saturation over segmented tissue regions in HSV space, where higher hue values correspond to reddish tones and lower values to blueish tones, and saturation reflects color intensity (Methods).

Mean (hue, saturation) values were (322.7, 0.153) for Munich, (313.9, 0.118) for Heidelberg, (311.3, 0.400) for Karolinska and (334.1, 0.270) for Radboud. Saturation varied most: Karolinska slides were 2.6-fold more saturated than the Munich training data and 3.4-fold more than Heidelberg, placing the external cohorts on opposite sides of the training distribution rather than displacing it in a single direction. Hue spanned 22.8 degrees, with Radboud shifted towards red and Karolinska towards blue-purple relative to Munich. 

Despite the diversity of their pretraining data, pathology foundation models retain sensitivity to scanner-induced appearance shift~\cite{de2025current, komen2026towards}. We therefore trained the report generation model with a dual-stage augmentation pipeline (supplementary material Fig. S1b and S1c). Each training tile was rendered in two additional color variants spanning a plausible range of H\&E appearances, one lighter with reduced contrast and saturation and one darker with increased contrast and saturation. In latent space, a lightweight auxiliary model predicted the embedding a patch would have received under a given augmentation, exposing the model to a differently augmented representation of every slide in every epoch at negligible computational cost. Unless stated otherwise, all reported results were obtained with this pipeline. Its contribution is isolated by ablation at the end of the following section.



\subsection*{Generalization to external cohorts}

Robust generalization across the scanner- and protocol-induced shifts described above is a prerequisite for clinical deployment. We therefore evaluated the report generation model on the three external cohorts without any fine-tuning or recalibration, using the same evaluation setup as for the internal cohort by extracting the relevant information from the generated reports. Results for the external cohorts are summarized in Table~\ref{tab:external_results}.


On the Heidelberg cohort our report generation model predicted the malignancy correctly for 93.0\% of all cases. The recall is 90.1\% while the precision is 93.7\% maintaining a strong F1-score of 91.9\%. Regarding grading, our report model achieved an accuracy of 67.9\% with an F1-score of 41.3\%. Compared to the internal test cohort the capability to determine malignancy of biopsies decreased slightly. Errors in grading remain dominated by adjacent grade misclassifications: for only 16 biopsies of the 164 correctly predicted malignant biopsies the Gleason score was off by more than one grade. Notably, none of the biopsies with a Gleason score of 7a or higher was predicted as benign.

For the Radboud and Karolinska cohorts our report generation model scored an accuracy of 92.6\% and 93.7\% respectively. The F1-scores for these cohorts are 95.0\% for Radboud and 95.4\% for Karolinska being only a small drop compared to the internal test cohort demonstrating strong generalization performance.
The grading F1-scores for the Radboud and Karolinska cohorts are 46.2\% and 50.1\%, respectively. As on the internal cohort, errors are dominated by adjacent-grade misclassifications, with the highest confusion for Gleason scores 6 and $\geq9$.
Across all external cohorts, our latent space augmentation strategy is consistently beneficial in mitigating the scanner-induced distribution shift and preserving diagnostic accuracy.

\begin{figure}[htbp]
\centering
\includegraphics[width=0.975\linewidth]{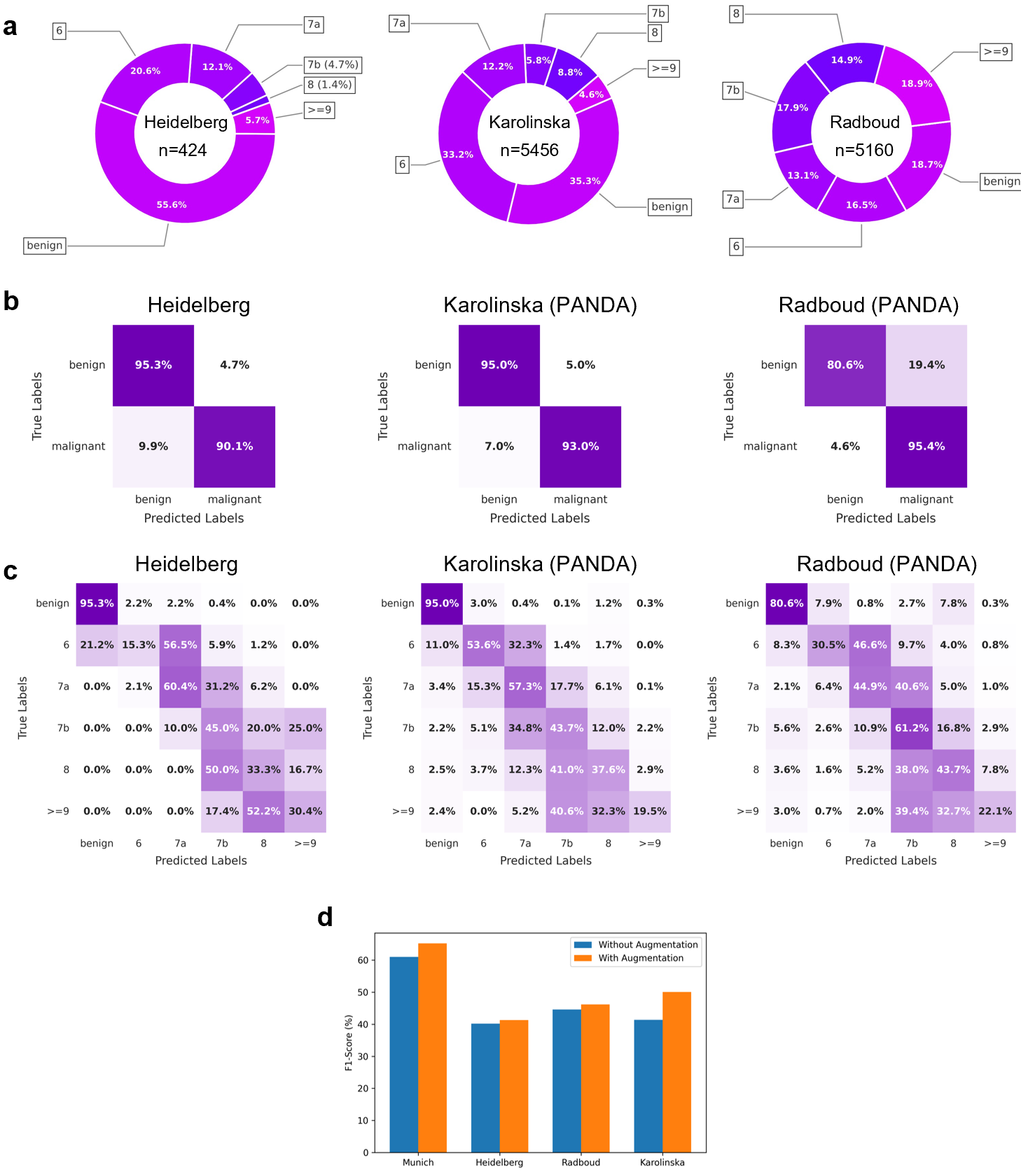}
\caption{\textbf{Evaluation of the report generation model on external test cohorts.} \textbf{a} Grade distribution across the three external test cohorts. \textbf{b} Confusion matrices for the malignancy status extracted from the generated reports. \textbf{c} Confusion matrices for the predicted Gleason grades per biopsy. \textbf{d} Our augmentation strategy consistently improves the model's ability to distinguish fine-grained differences between Gleason scores, as demonstrated by an increase in the macro F1-score.}
\label{fig:result_external}
\end{figure}

To isolate the contribution of the augmentation pipeline, we compared these results against an otherwise identical model trained without augmentation. Augmentation improved grading F1-score on every cohort, by +8.7 percentage points on Karolinska, +1.6 on Radboud and +1.5 on Heidelberg (Fig.~\ref{fig:result_external}). Performance also improved on the internal Munich cohort (+4.2 percentage points), where no domain shift is present by construction, indicating that the augmentation regularizes slide-level aggregation as well as improving cross-cohort transfer.

\begin{table}[ht]
\centering
\begin{tabular}{lcccccccc}
\toprule
 & \multicolumn{4}{c}{\textbf{\textls[100]{Malignancy}}} & \multicolumn{4}{c}{\textbf{\textls[100]{Grading}}} \\
\cmidrule(lr){2-5}
\cmidrule(lr){6-9}
\textbf{Cohort} & Accuracy & Precision & Recall & F1-score & Accuracy & Precision & Recall & F1-score \\
\midrule
Heidelberg & 93.0 & 93.7 & 90.1 & 91.9 & 67.9 & 47.1 & 46.6 & 41.3 \\
Karolinska & 93.7 & 97.2 & 93.0 & 95.0 & 65.1 & 56.4 & 51.1 & 50.1 \\
Radboud & 92.6 & 95.5 & 95.4 & 95.4 & 47.7 & 52.4 & 47.2 & 46.2 \\
\bottomrule
\end{tabular}
\caption{\label{tab:external_results}Accuracy, Precision, Recall and F1-score for malignancy and grading of the report generation model on the external cohorts using the described augmentation strategy.}
\end{table}

\section*{Discussion}
We show that a report generation model trained end-to-end from scratch, without any pretrained text encoder, learns the clinically relevant information contained in pathology report text. On our internal test cohort, malignancy and grading performance was comparable to that of a classification model trained directly on extracted labels, indicating that the additional burden of generating free text does not compromise diagnostic accuracy.

Robustness and generalization are central challenges when transferring AI models for pathology into clinical practice. We incorporated a dual strategy of deterministic, realistic color augmentations and latent-space augmentations, which yielded consistent grading improvements across our internal and three external cohorts. This strategy is transferable to any training pipeline in which foundation-model embeddings serve as input, and could therefore benefit models for other organs, particularly in the low-data regime where only few slides are available.

Across all cohorts, grading errors were overwhelmingly adjacent-grade rather than gross misclassifications, a pattern that mirrors the inter-observer variability well documented in prostate grading. How this variability, present in reports authored by multiple pathologists over nearly a decade, propagates to model training and evaluation remains an open question that warrants further study.
The color augmentations were tuned for realistic appearance and merit more systematic evaluation, though generating multiple embedding variants per WSI does increase computational and storage requirements.

On our internal test cohort, our model reaches malignancy-detection accuracy comparable to that of a clinically approved tool, despite a substantially smaller and uncurated training corpus. This comparison favors our model insofar as it was trained on data from the same institution and scanner, whereas the approved tool was developed independently. It nonetheless demonstrates that individual institutions can develop a report generation model tailored to their own patient population and reporting language with manageable effort.

Unlike classification models that output a probability distribution, our framework produces a native-language draft report that can be integrated directly into the pathologist's reporting workflow. This positions the model as an assistive drafting tool rather than an autonomous decision-maker: the pathologist reviews, corrects, and signs off on each generated text, retaining full diagnostic responsibility. Such a human-in-the-loop deployment aligns with how generative tools are most safely introduced into clinical practice, where the model reduces documentation burden while the expert remains in control of the final diagnosis. A generative model does, however, introduce a failure mode absent in classifiers: a confidently phrased but incorrect grade or infiltration value embedded in otherwise fluent text. Our use of deterministic greedy decoding makes generated reports reproducible and thus auditable, but it does not eliminate this risk. Consequently, the safe use of such a system rests on the assumption that generated drafts are always verified against the underlying slide, and this requirement should be made explicit in any deployment protocol.

Our study has limitations.
First, model performance is weakest for the rare high-grade classes, most notably Gleason score $\geq9$, where the scarcity of training examples limits the model's ability to recognize the corresponding morphologic patterns. Targeted enrichment of these underrepresented grades, which our automated pipeline makes feasible at low cost, is a clear path to improvement.
Second, although the model learns to estimate tumor infiltration and to distinguish the two measurement protocols, the absolute errors (mean absolute errors of 2.9 mm for extent and 15.2\% for percentage, with $R^2$ scores below 0.5) are too large for the predicted values to serve as precise clinical measurements.
The predicted infiltration should therefore be interpreted as an approximate estimate of tumor burden that is refined by the pathologist's own quantification, consistent with an assistive, human-in-the-loop role of the system.
Finally, our clinical-attribute evaluation relies on a large language model to extract attributes from both the original and the generated reports. Our manual audit validated this extraction on human-authored reports, but generated text may deviate in phrasing or completeness in ways the audit did not sample, and a systematic bias in the extractor would affect both sides of the comparison rather than cancelling out. Direct pathologist review of generated reports would provide a stronger validation of text quality than attribute extraction alone.

Our evaluation is retrospective and based entirely on archival data. A future research direction is to conduct a prospective or reader study that is necessary to establish clinical deployment and to quantify the model's effect on pathologist workload and accuracy in practice.


\section*{Methods}

\subsection*{Ethics statement}
This study was approved by the Ethics Committee of the Technical University of Munich (approval number 2023-665-1-S-CT) and was conducted in accordance with the Declaration of Helsinki. Given the retrospective design and the exclusive use of archived material without patient participation, the requirement for written informed consent was waived. All data used in this study was anonymized prior to model training.

\subsection*{Datasets}

\subsubsection*{Munich cohort}
All digitally available H\&E-stained prostate biopsy WSIs from the Institute of Pathology of the Technical University of Munich were collected resulting in a dataset of 17,344 WSIs from 2,402 cases spanning the years from 2016 to 2026. All WSIs were scanned with a Leica Aperio GT 450 DX scanner with a magnification of 40x at 0.26 micrometers per pixel. The dataset was split into train and test partitions stratified by case and availability of predictions by the FDA-cleared AI tool used in the evaluations. Predictions for this tool are only available for the test partition. During the training stage of the model the train partition was further split into train and validation partitions stratified by case. By utilizing historical archive data, the dataset’s morphological distribution is representative of real-world clinical prevalence. The pathology reports corresponding to the cases are in German language and were obtained from the Institute's LIS. No translations were made for this study. The reports were anonymized removing doctor's names and patient information before further processing.

\subsubsection*{Heidelberg cohort}
The 424 prostate biopsy samples from 50 cases of the Heidelberg cohort were processed, stained and scanned at the Institute of Pathology at Heidelberg University. Selection of the cases was done randomly with the only premise of containing at least a single malignant biopsy core ensuring that the cohort reflects clinical reality. All biopsies were scanned using a Leica Aperio GT 450 (research use only) scanner with a magnification of 40x at 0.26 micrometers per pixel. Ground truth labels on malignancy and grading for evaluation were extracted from the  corresponding pathology reports.

\subsubsection*{PANDA (Karolinska and Radboud cohort)}
The dataset of the PANDA challenge~\cite{bulten2022artificial} is the largest publicly available dataset for WSIs of prostate biopsies. It comprises cohorts from the Karolinska Institutet in Sweden and the Radboud University Medical center in the Netherlands. It includes labels on malignancy and Gleason grading.

The Karolinska cohort consists of 5,456 prostate biopsy samples scanned with an Aperio AT2 scanner at 0.45 and 0.5 micrometers per pixel. Ground truth information was determined by one uropathologist.

The Radboud cohort consists of 5,160 prostate biopsy samples scanned with a 3DHISTECH Pannoramic Flash II 250 scanner at 0.24 micrometers per pixel downsampled to 0.48 micrometers per pixel.

\subsection*{Creation of WSI-text pairs and evaluation labels}
The reports of the Munich cohort are further processed to create WSI-text pairs and labels for training the report generation model and the evaluation. The reports of the Munich cohort describe all specimens of a case in a single document, so one main task to obtain WSI-text pairs was the disaggregation of these composite reports into biopsy-specific descriptions (Fig.~\ref{fig:llm_extraction}). This text processing was performed with the LLM gpt-oss-120b \cite{openai2025gptoss120bgptoss20bmodel}, deployed locally within the secure institutional environment so that no report data left the hospital network. During LLM inference, the parameters \textit{temperature} and \textit{top-p} were set to 0.1 and 0.9 respectively. The LLM was run with 4-bit quantization. Extraction steps were performed using a few-shot learning paradigm. In preliminary experiments, this reduced hallucinated content and improved adherence to the specimen numbering conventions used in the LIS compared with zero-shot prompting.

Each report was first segmented into its two primary sections, microscopy and critical findings. Each section was processed separately. Within each section, extraction was guided by the specimen numbers associated with each WSI: the model was prompted to isolate the text segments corresponding to a given specimen identifier, yielding a microscopy and a critical findings text for every biopsy core. These texts were concatenated and paired with the WSI of the corresponding biopsy to obtain the WSI-text pair.

To enable an evaluation that captures diagnostic correctness rather than linguistic similarity alone (see Evaluation Metrics), the same LLM was used to extract structured ground-truth labels for each biopsy from the corresponding specimen-level text: malignancy status (benign vs.\ malignant), Gleason grade, and infiltration percentage or extent. These parameters correspond to the values required for prostate cancer diagnosis and treatment planning \cite{van20202019}.

The accuracy of the pipeline was assessed on 200 biopsy cores sampled at random from the full cohort. For each sampled biopsy, the extracted labels were compared against the original reports in the LIS by a trained annotator. A discrepancy was defined as any mismatch in an extracted label.

\subsection*{WSI preprocessing}
A WSI of a prostate biopsy can contain several consecutive sections from the same tissue block and span more than
$100{,}000\times100{,}000$ pixels, which is too large to process as a single image. Each WSI was therefore tessellated into patches using the
TRIDENT~\cite{zhang2025standardizing} toolkit, which builds on the
GrandQC~\cite{weng2024grandqc} segmentation model. Tissue was first segmented from irrelevant regions such as background and pen marks. The segmented tissue was then divided into non-overlapping patches of $512\times512$ pixels at 0.26 micrometers per pixel. Each patch was downsampled to $224\times224$ pixels to match the input size of
the pathology foundation model UNI2-h~\cite{chen2024towards}, resulting in an effective resolution of 0.59 micrometers per pixel. RGB values were scaled to the range $[0,1]$ and normalized using the ImageNet mean and standard deviation. UNI2-h was kept frozen and produced a 1536-dimensional embedding for each patch.

\subsection*{Text tokenizer}
Because the report generation model was trained from scratch without pre-trained components, we trained a domain-specific Byte-Pair Encoding (BPE) tokenizer~\cite{sennrich2016neural} on the German report texts using the Hugging Face \texttt{tokenizers} library~\cite{wolf2019huggingface}. The corpus comprised the texts of all WSI-text pairs in the training split. Normalization Form Canonical Composition (NFC) was applied while preserving casing and umlauts, and numbers were split into
individual digits to represent clinical measurements consistently. Requiring a minimum merge frequency of two yielded a vocabulary of 10,142 tokens. 

Each sample was additionally structured with control tokens. The microscopy and critical findings sections were enclosed in dedicated section markers so that the generated reports retain the structure of the original reports. A token encoding the Gleason score was prepended to the report text: because generation proceeds autoregressively, the model predicts this token first at inference and conditions the remainder of the report on its own grade prediction.

\subsection*{Model architecture}
The report generation model follows the CoCa framework~\cite{yucoca} and
comprises a slide encoder, a text encoder and a multimodal decoder.
The frozen foundation model UNI2-h outputs a 1536-dimensional embedding per patch. A trainable MLP projects each embedding to 768 dimensions before it enters the slide encoder, which consists of six transformer blocks~\cite{vaswani2017attention} with 12 attention heads. A trainable CLS token is prepended to the patch sequence and serves as the slide-level representation. Because the number of patches varies between slides, sequences were zero-padded within a batch and padded positions
were excluded from attention by an additive mask.

To incorporate spatial relationships between patches, we adapted ALiBi
positional encoding~\cite{presstrain} to two dimensions. Each patch $p_i$ was assigned its coordinates $(x_{p_i}, y_{p_i})$ on the slide, scaled to $[0,1]$ per slide, and the Euclidean distance between two patches was computed as
\begin{equation}
    d(p_1, p_2) = \sqrt{(x_{p_1} - x_{p_2})^2 + (y_{p_1} - y_{p_2})^2}.
    \label{eq:alibi_dist}
\end{equation}
This distance was subtracted from the pre-softmax attention logits of head $h$, scaled by a head-specific slope $m_h$ and a global factor $\gamma$:
\begin{equation}
    \mathrm{attn}_h(p_1, p_2) =
    \frac{q_{p_1}^{\top} k_{p_2}}{\sqrt{d_k}} - \gamma \, m_h \, d(p_1, p_2).
    \label{eq:alibi_bias}
\end{equation}
The head-specific slopes $m_h$ follow the geometric schedule of~\cite{presstrain}, including the interpolation scheme used when the number of heads is not a power of two, and were kept fixed during training. Unlike in the original one-dimensional formulation, the penalty is symmetric and applied bidirectionally, as attention over patches is not causally masked. The CLS token was exempted from the penalty by setting its distance to all patches to zero, so that the slide-level representation aggregates patch features without spatial bias. Nearby patches are thereby favored over distant ones, encoding slide geometry without learned positional embeddings.

The text encoder consists of 12 transformer blocks with an embedding dimension of 768 and 12 attention heads. It processes the token sequence produced by the tokenizer with causal masking, so each token attends only to preceding tokens. A CLS token appended to the end of the sequence yields the text-level representation used for the contrastive objective.

The multimodal decoder consists of 12 transformer blocks that receive the output of the unimodal decoder and attend to the slide representation through cross-attention, with text embeddings as queries and the encoded patch embeddings as keys and values. Its output is projected onto the tokenizer vocabulary to yield a probability distribution over the next token.

The model was trained with the CoCa objective, a weighted sum of a contrastive and a captioning loss,
\begin{equation}
    \mathcal{L} = \lambda_{\mathrm{con}} \mathcal{L}_{\mathrm{con}}
                + \lambda_{\mathrm{cap}} \mathcal{L}_{\mathrm{cap}},
\end{equation}
with $\lambda_{\mathrm{con}} = \lambda_{\mathrm{cap}} = 1$.
$\mathcal{L}_{\mathrm{con}}$ is the symmetric InfoNCE loss~\cite{radford2021learning} between the slide and text CLS tokens over all pairs in a batch, with a learnable temperature. $\mathcal{L}_{\mathrm{cap}}$ is the token-level cross-entropy maximizing the likelihood of each token given all preceding tokens and the slide representation.

\subsection*{Text generation}

During inference, textual reports are generated autoregressively starting with an empty text sequence. While standard LLMs often employ stochastic sampling techniques for selection of the next token (e.g. top-$k$~\cite{fan2018hierarchical} or top-$p$~\cite{holtzman2019curious}) to increase variety, greedy sampling ($k=1$) is used to generate our report texts. This deterministic approach ensures that the model always selects the token with the highest predicted probability. In a diagnostic context, this is crucial. For instance, when predicting Gleason grades or infiltration percentages, the model must prioritize clinical precision over linguistic variety, as even minor deviations in token selection could lead to considerable diagnostic discrepancies.

\subsection*{Quantification of color domain shift}

To characterize appearance differences between cohorts, 100 WSIs were drawn at random without replacement from each of the Munich, Heidelberg, Karolinska and Radboud datasets. For each slide, tissue was segmented from background using the segmentation results from the WSI preprocessing at thumbnail resolution. Pixels were converted from RGB to HSV space using
scikit-image~\cite{van2014scikit}, with hue expressed in degrees on
$[0, 360)$ and saturation on $[0, 1]$. Because hue is a circular quantity, per-cohort mean hue $\bar{h}$ was computed
using circular statistics over the $N$ pooled tissue pixels. Saturation was averaged arithmetically.

\subsection*{Augmentation pipeline}

Models trained on single-institution, single-scanner data transfer poorly to external cohorts, and pathology foundation models retain measurable sensitivity to scanner-induced appearance shift despite the diversity of their pretraining corpora\cite{de2025current, komen2026towards, thiringer2026scanner}. We therefore augmented the training data at two levels: at the pixel level prior to feature extraction, and in the latent space of the frozen patch encoder during report generation training.

\subsubsection*{Pixel-level color augmentation}

For every WSI in the Munich training set, two additional color variants were generated alongside the original WSI by applying deterministic
transformations to each extracted tissue tile prior to feature extraction: a lighter variant with brightness and saturation factors of 1.2 and 0.8, and a darker variant with gamma and saturation factors of 1.35 and 1.35 (implemented with torchvision). Transformation parameters were chosen empirically on the Munich dataset to span a plausible range of H\&E appearances without producing visually implausible tissue. No cohort-specific tuning and no data from the external cohorts were used at any point.

Each of the three variants was passed independently through the frozen patch encoder, yielding three embedding sets per slide. During
report generation training, one of the three sets was sampled uniformly at
random per slide per epoch. Validation and test slides were never augmented and were processed only in their unmodified form.

\subsubsection*{Latent space augmentation}
Repeatedly passing augmented images through a large foundation model is
computationally prohibitive, so additional augmentation was applied directly in embedding space, following the approach of HistAug\cite{boutaj2025controllable}. An auxiliary model consisting of a single transformer block was trained to predict the embedding of an augmented patch given the embedding of the corresponding unaugmented patch and a vector encoding the target augmentation parameters. The parameter vector covered rotation, flipping, Gaussian blur, hue, saturation, contrast, and brightness.

The auxiliary model was trained on a single Nvidia RTX 3090 GPU on 3.8 million patches randomly drawn from slides of the Munich training split only, using MSE loss between predicted and ground-truth augmented embeddings, optimizer AdamW with learning rate $5\times10^{-4}$, batch size 32. It was then frozen. During report generation training the augmentation model was run in inference mode with randomly sampled augmentation parameters per slide per epoch, so that the downstream model encountered a distinct latent representation of each slide in every epoch. The auxiliary model was not applied at validation or test time.

\subsection*{Model training setup}
\subsubsection*{Report generation model}
All weights are randomly initialized. Following the initialization strategy of CLIP~\cite{radford2021learning}, the standard deviation of the residual projection layers is scaled down by the network depth to ensure stable convergence.

To manage GPU memory consumption during training, a maximum of 4,800 patches is dynamically sampled from each WSI. For the majority of WSIs in our dataset, this threshold captures all available patches. Because batches must be zero-padded to match the longest sequence in that specific batch, the extreme padding-induced memory overhead is mitigated by truncating the outlier WSIs that exceed this maximum sequence length.

As optimizer AdamW~\cite{loshchilov2017decoupled} was used with betas of (0.9, 0.999) and a weight decay of 0.2. The learning rate schedule used a warm-up period up to 10,000 iterations with a learning rate of $5\times10^{-5}$ and then cosine annealed to 0 over a period of 200 epochs. The batch size is set to 32 and the training is run on NVIDIA H100 GPUs.

To overcome the imbalance of grade distribution in our training dataset, oversampling based on the metadata obtained during the extraction stage is employed. Each sample has a probability inversely proportional to its grade to be sampled during the training stage. In this way, the model has the possibility to encounter less frequent morphologic patterns more often.

\subsubsection*{ABMIL model}
The weights of the ABMIL model were randomly initialized using Glorot initialization~\cite{glorot2010understanding}. An AdamW optimizer is used with betas of (0.9, 0.999) and a learning rate of 0.01. The learning rate was divided by 10 if the accuracy did not improve for five epochs. The batch size is set to 32 and the training is run on an NVIDIA H100. Weighted Cross-Entropy is used as loss to train the model to overcome the imbalance of Gleason scores in the dataset.

\subsection*{Attention maps}

The slide encoder uses the transformer architecture making use of the self-attention mechanism. Inside each block of the transformer, attention values are computed for each patch embedding processed by the slide encoder. These attention values should correspond to the importance of the respective patch and provide explainability which areas of the slide have influenced the generated text the most.

Attention values are scattered across all blocks of the transformer. To obtain a single attention value per patch from the WSI, the attention-rollout aggregation method is used. It is a heuristic method that recursively multiplies the attention scores for each patch to capture the information flow with respect to the attention. The resulting attention values per patch can then together with the original coordinates be used to highlight important areas.

\subsection*{Evaluation metrics}
Evaluating the quality of generated pathology report texts is a complex task. Standard natural language processing (NLP) metrics, such as BLEU, ROUGE, and METEOR, are commonly used for this purpose. However, because they originated in the domains of machine translation and text summarization, their ability to capture the clinical semantics of generated text is limited.
Moreover, the metrics may be artificially inflated by the presence of common repetitive phrases that carry no considerable diagnostic value.

To address these limitations, we propose an evaluation protocol focused on core clinical attributes that, according to local guidelines for prostate cancer~\cite{LeitlinieProstatakarzinom2025}, must be present in a biopsy report. Specifically, we focus on malignancy, grading (Gleason score), and infiltration. For the PANDA dataset cohorts, ground-truth labels for malignancy and grading are natively provided. For our internal dataset, a ground truth is established by using a Large Language Model (LLM) to extract these key attributes from the original, human-authored pathology reports. To evaluate the generated reports, the identical LLM extraction method is applied to isolate the same attributes from the newly generated texts.

The values extracted from the generated reports are then directly compared against the ground truth. Malignancy and grading are evaluated as categorical variables. We compute standard classification metrics, including accuracy, precision, recall, and F1-score, utilizing macro-averaging for the latter three to account for class imbalance. Infiltration, conversely, is a continuous variable recorded either as a percentage or an absolute length, depending on the measurement protocol used which changed over time in our dataset. Results for these two measurement protocols are reported separately. To evaluate infiltration accuracy, Pearson's correlation coefficient is calculated to assess the overall trend, alongside Median Absolute Error to quantify absolute deviation.

While traditional n-gram-based NLP metrics emphasize syntactic similarity and overall phrasing, our evaluation protocol prioritizes the factual correctness of critical clinical information. In the medical domain, where a single omitted word or altered number can fundamentally change a diagnosis, we believe this attribute-centric approach provides a more robust and clinically relevant measure of model performance.

\bibliography{sample}

\ifsupp
\clearpage
\beginsupplement
\begin{center}
  {\Large\bfseries\sffamily\ Supplementary Material}
\end{center}

\lstset{basicstyle=\ttfamily\footnotesize, breaklines=true,breakindent=0pt, frame=single, columns=fullflexible, xleftmargin=0pt}
\begin{lstlisting}[caption={Prompt for extracting sections from the report. The Output can automatically be parsed to split the sections. Prompt was translated from German to English for readability.},captionpos=b, label={lst:prompt_sections}]
Extract the sections Clinical Information/Question, Macroscopy, Microscopy and Critical Report from the following pathology report. If there are one or more additional reports, the information belonging to these sections shall be merged into a single section. The sections must be labeled with the respective headings "Clinical Information/Question:", "Macroscopy:", "Microscopy:" and "Critical Report:", if present. The output shall be produced without any introduction or closing remarks. If a section is not present in the report, it shall be omitted from the output. Personal data and contact information shall be removed. The report text must not be modified.

Pathology report:
{report}
\end{lstlisting}

\lstset{basicstyle=\ttfamily\footnotesize, breaklines=true,breakindent=0pt, frame=single, columns=fullflexible, xleftmargin=0pt}
\begin{lstlisting}[caption={Prompt for extracting specimen-level information from a section. Prompt was translated from German to English for readability.},captionpos=b,label={lst:prompt_specimen}]
The following text of a pathology report contains descriptions of several tissue blocks. Each tissue block is identified by a number (e.g. 1., 2., 3.). Descriptions may also refer to multiple tissue blocks (e.g. 1.-3. or 1., 2.). Your task is to extract all text passages that refer to the specified tissue block. Output only the extracted text without any additional explanation. If no information on the specified tissue block is available, output only "No information".

Tissue block: {block}.

Pathology report:
{report}
\end{lstlisting}

\lstset{basicstyle=\ttfamily\footnotesize, breaklines=true,breakindent=0pt, frame=single, columns=fullflexible, xleftmargin=0pt}
\begin{lstlisting}[caption={Prompt for extracting labels (here Gleason score) from a specimen description. Prompt was translated from German to English for readability.},captionpos=b,label={lst:prompt_label}]
Extract the Gleason score from the following prostate pathology report. Output only the Gleason score. If the tissue specimen is free of carcinoma, output only "no carcinoma". If no information is available, output only the following text: "No information".

Pathology report:
{report}
\end{lstlisting}

\begin{figure}[htbp]
\centering
\includegraphics[width=0.9\linewidth]{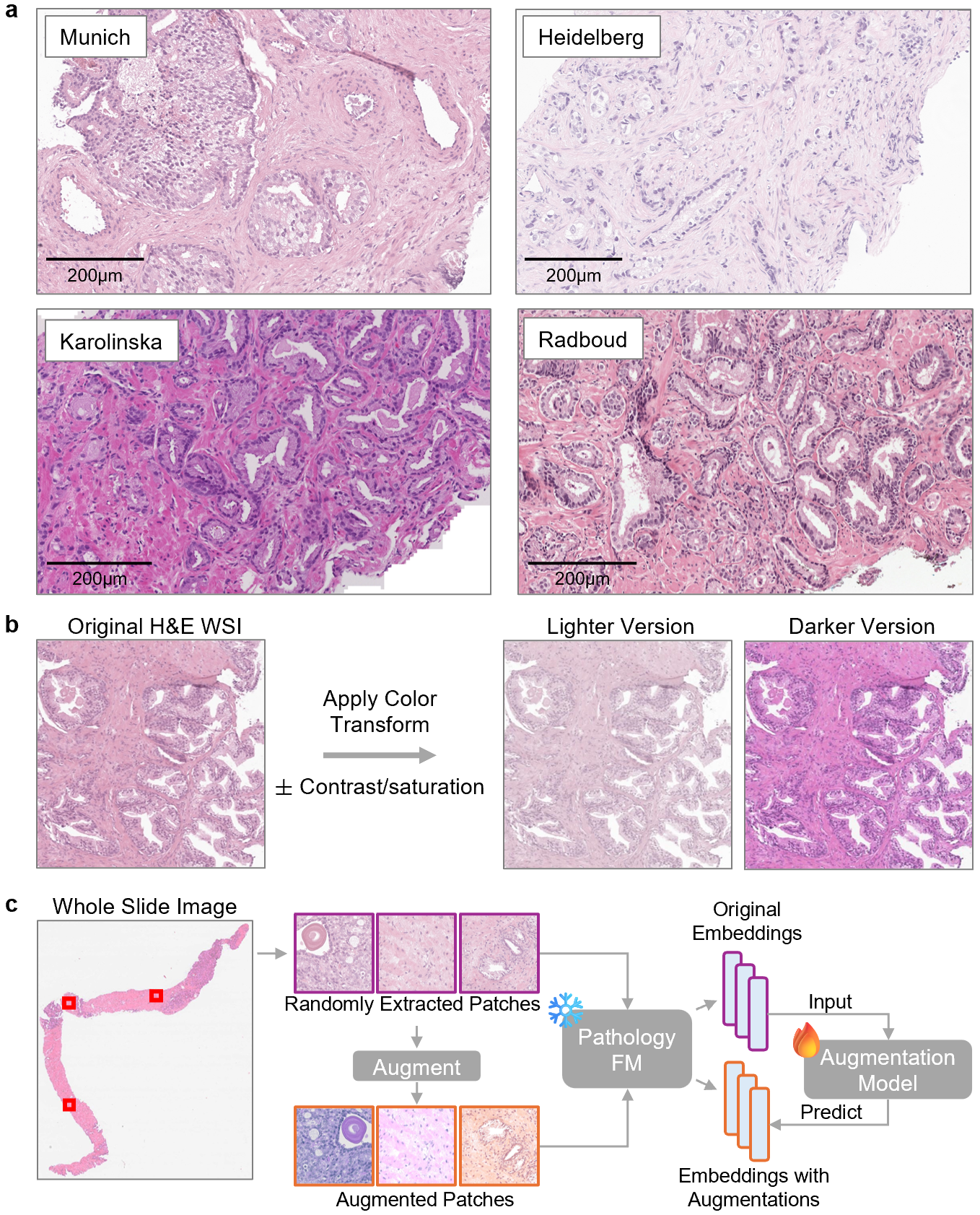}
\caption{\textbf{Overview of color variations and augmentation strategy.} \textbf{a} H\&E-stained WSIs exhibit substantial visual variability depending on the scanner and staining protocol. Sample images from each cohort used in this study highlight this variance. \textbf{b} For every tissue patch extracted for training, two additional color variants are generated: a lighter version with reduced contrast and saturation, and a darker version with increased contrast and saturation. This approach aims to simulate realistic staining variations encountered in clinical practice. \textbf{c} To further diversify the training data, a latent space augmentation strategy is applied. We train an auxiliary augmentation model on a subset of the training images to predict the embedding of an augmented patch, given the original non-augmented patch embedding and the target augmentation parameters.}
\label{fig:augmentation}
\end{figure}
\fi

\end{document}